\documentclass[a4paper,preprint,1p,11pt]{elsarticle}
\usepackage{lineno,hyperref}
\usepackage{amssymb}
\usepackage{algorithm}
\usepackage{algorithmic}
\usepackage{longtable}
\usepackage{rotating}
\usepackage{blindtext}
\usepackage{multirow}
\usepackage{amsmath}
\usepackage{amssymb}

\usepackage{hyperref}
\hypersetup{
    colorlinks=true,
    linkcolor=black,
    filecolor=magenta,      
    urlcolor=black,
    citecolor=black
}

\journal{Computers \& Operations Research}

\begin{document}

\begin{frontmatter}

\title{Experimental Analysis of Design Elements of Scalarizing Functions-based Multiobjective Evolutionary Algorithms}

\author{Mansoureh Aghabeig}
\ead{mansoureh.aghabeig@yahoo.com}
\author{Andrzej Jaszkiewicz}
\address {Poznan University of Technology, Faculty of Computing, Institute of Computing Science, ul. Piotrowo
2, 60-965 Poznan}
\ead{andrzej.jaszkiewicz@put.poznan.pl}

\begin{abstract}
In this paper we systematically study the importance, i.e., the influence on performance, of the main design elements that differentiate scalarizing functions-based multiobjective evolutionary algorithms (MOEAs). This class of MOEAs includes Multiobjecitve Genetic Local Search (MOGLS) and Multiobjective Evolutionary Algorithm Based on Decomposition (MOEA/D) and proved to be very successful in multiple computational experiments and practical applications. The two algorithms share the same common structure and differ only in two main aspects. Using three different multiobjective combinatorial optimization problems, i.e., the multiobjective symmetric traveling salesperson problem, the traveling salesperson problem with profits, and the multiobjective set covering problem, we show that the main differentiating design element is the mechanism for parent selection, while the selection of weight vectors, either random or uniformly distributed, is practically negligible if the number of uniform weight vectors is sufficiently large.
\end{abstract}


\end{frontmatter}

\section{Introduction}

In many areas, an optimal decision should take into consideration two or more conflicting objectives. The multiobjective optimization problem (MOOP) is an optimization problem which involves multiple objective functions and in mathematical terms can be formulated as:
\[ ``{\textrm{minimize}} \textrm{''}[f_1(x)=z_1,...,f_J(x))=z_J]\]
\[s.t. \quad x \in D\]
where \textit{solution} $x=[x_1,...,x_I]$ is a vector of decision variables, \(D\) is the set of feasible solutions.

The image of a solution \(x\) in the objective space is a \textit{point} \(z^x = [z_1^x,...,z_J^x] = f(x)\) such that \(z_j^x = f_j(x),\quad j = 1,...,J\).

Point $z^1$ dominates $z^2$, $z^1 \succ z^2 $,  if $\forall_j \quad  z_j^1 \leq z_j^2$ and $z_j^1 < z_j^2$ for at least one $j$. The solution is a \textit{Pareto  solution}, if there does not exist another feasible solution which dominates it. The image of a Pareto solution in the objective space is called a \textit{non-dominated point}. The set of all Pareto solutions is called the \textit{Pareto set}, and the image of the Pareto set in the objective space is called the \textit{non-dominated frontier} or the \textit{Pareto front}. Two solutions are \textit{mutually non-dominated} if nither of them dominates the other and their images in the objective space are different. In this paper, a set of mutually non-dominated solutions generated by a multiobjective evolutionary algorithm is called a \textit{Pareto archive}.

A MOOP is called a \textit{multiobjective combinatorial optimization problem}, if it has two characteristics. First, the decision variables are discrete, and second, the set of feasible solutions is finite. Combinatorial optimization finds applications in many real world problems, such as scheduling, time tabling, production, facilities design, routing, and many others \cite{yu2013industrial}.

In multiobjective optimization, there is usually no single solution which simultaneously achieves the optimum values for all objectives, making the preferences of a decision maker (DM) necessary for selecting the most preferred solution. It is a generally accepted assumption that the DM's preferences are compatible with the dominance relation. Under this assumption, the most preferred solution belongs to the Pareto set. Thus, the goal of most multiobjective optimization algorithms is finding the Pareto set, or a good approximation of it, for further exploration by the DM. 

Evolutionary algorithms are a promising option for solving MOOPs \cite{deb2001multi}. They process a population of candidate solutions in each iteration, so they are able to search for multiple (approximately) Pareto solutions concurrently in a single run.

In evolutionary algorithms, in each intermediate iteration,
a selection process is performed in which good members
of the population have a higher probability to survive and
worse members have a higher probability to be eliminated.
Thus, an evaluation mechanism of intermediate solutions is
necessary. In the single objective case, intermediate solutions
can naturally be evaluated with the value of the single objective
function. In the multiobjective case, there is no such obvious
evaluation mechanism, so different mechanisms are used in
different methods.

An often used type of evaluation mechanism is the Pareto dominance-based evaluation, which uses the dominance relation to evaluate solutions. A typical example is Pareto ranking used in the NSGA2 algorithm \cite{deb2002fast}; PAES \cite{knowles1999pareto} and SPEA2 \cite{zitzler1999multiobjective} are examples of other methods using this type of mechanism. Although Pareto dominance-based mechanisms have the advantage of not requiring the transformation of a MOOP into a single objective problem, they may suffer from other weaknesses, such as losing selection pressure with the increasing number of objectives, needing an additional mechanism for preserving diversity, and their hybridization with local search not being straightforward \cite{neri2012handbook}.
Another type of evaluation mechanism is scalarizing function-based evaluation. In  mechanisms of this type, a multiobjective optimization problem is transformed into a family of parametric single objective optimization problems. In each problem, a non-negative weight vector defines a single scalarizing function. Two typically used scalarizing functions are given below.

\textit{Weighted linear scalarizing functions} are defined in the following way:\[s_1(z,\Lambda) = \sum_{j=1}^{J} \lambda_iz_j\]
where $ \Lambda = [\lambda_1,...,\lambda_J] \mbox{ } \forall  \lambda_j \geq 0,$ is a weight vector. 
\\Each weighted linear scalarizing function has at least one global optimum belonging to the Pareto set  \cite{Steuer1985MCO}.

\textit{Weighted Chebycheff scalarizing functions} are defined in the following way:
\[s_\infty(z,z^*,\Lambda)= -max_j(\lambda_j(z_j^*-z_J))\]
where $ z^*$ is a reference point, $ \Lambda = [\lambda_1,...,\lambda_j]\mbox{ }\forall  \lambda_j \geq 0,$ is a weight vector. 
\\Each weighted Chebycheff scalarizing function has at least one global optimum belonging to the Pareto set. For each Pareto solution $x$, there exists a weighted Chebycheff scalarizing function such that $x$ is a global optimum of $s_\infty$ \cite{Steuer1985MCO}.  

The above properties suggest an advantage of weighted Chebycheff scalarizing functions, since every Pareto solution may be obtained by optimizing this type of function. However, this property holds only if an exact optimum solution of the function can be obtained. In practice, when heuristic methods are used, linear scalarizing functions often perform better \citep{zhang2007moea,jaszkiewicz2002genetic}.

The two classes of functions can also be combined, producing \textit{mixed scalarizing functions}, defined in the following way: 
\[s_{m}(z,z^*,\Lambda) = w_1s_1(z,\Lambda)+w_\infty s_\infty(z,z^*,\Lambda)\]
where $w_1$ defines the weight of the linear scalarizng function and $w_\infty$ defines the weight of the Chebycheff scalarizing function. The sum of these two weights should equal one.

Two well-known examples of multiobjective evolutionary algorithms based on scalarizing functions are MOGLS \cite{jaszkiewicz2002genetic} and MOEA/D \cite{zhang2007moea}. These methods proved to be very successful in multiple computational experiments \citep{gong2012community,ishibuchi2015behavior,ke2013moea,zhang2010expensive,li2014adaptive,ding2013modified,liu2014decomposition,jaszkiewicz2002IEEETEC,jaszkiewicz2003IEEETEC,mei2011decomposition} and practical applications \citep{sengupta2012evolutionary,sengupta2013multi,carvalho2012multi,trivedi2015enhanced}. 

Although the ways MOGLS and MOEA/D were presented in the original papers were very different, the two methods have a very similar structure. In fact, they differ only in two main elements: the selection of the weight vectors defining scalarizing functions, and the selection of parents for recombination. 

The main goal of this paper is to experimentally assess which of these two elements has greater influence on the performance of MOGLS and MOEA/D, and which version of each element results in better performance. 

The rest of this paper is organized as follows. In section \ref{II}, a short description of MOGLS and MOEA/D is given. An intermediate method between MOGLS and MOEA/D, called Uniform MOGLS (UMOGLS), is introduced in section \ref{III}. The computational experiments and discussion of the obtained results are presented in sections \ref{IV} and \ref{V} respectively. The paper ends with conclusions and potential directions for future research.

\section{MOGLS and MOEA/D algorithms}\label{II}
The main idea of scalarizing function-based multiobjective algorithms is as follows: if we optimized all weighted Chebycheff scalarizing functions defined by all possible weight vectors, we would obtain the true Pareto set. Unfortunately, implementing this idea in practice is impossible, since the set of all weight vectors is infinite, and, in many cases, there exists no exact method for finding the optimum solution of a scalarizing function within a realistic time frame. However, we can still approximate the Pareto set by the heuristic optimization of a set of various scalarizing functions defined by a set of well-distributed weight vectors. 

From another point of view, MOGLS and MOEA/D are based on the single objective Genetic Local Search (sGLS) algorithm. In each iteration of sGLS, two solutions (parents) are chosen for recombination from a population of solutions being relatively good on the objective function. In other words, the parents chosen for recombination are relatively good on the objective function. The offspring is generated by a recombination of the parents and then improved by a local search. 

A single iteration in both MOGLS and MOEA/D is almost the same as a single iteration of sGLS, i.e., in each iteration they select two solutions which are relatively good on the current scalarizing function, and the offspring is then improved by a local search guided by the same function. However, for each iteration, a different weight vector, and thus a different scalarizing function, is selected. Furthermore, the two algorithms use special mechanisms for parent selection. These mechanisms are a necessity, as the populations used in multiobjective algorithms are relatively large and contain solutions that are dispersed over various regions of the objective space, only some of them being good on the current scalarizing function (while with single objective algorithms all solutions in the population are usually relatively good on the single objective function). In short, MOGLS and MOEA/D use two specific mechanisms: one for choosing weight vectors, and another for choosing two parents which are relatively good on the current scalarizing function. The two mechanisms differ in each method. 

MOGLS was proposed by Jaszkiewicz \cite{jaszkiewicz2002genetic} and further developed in \cite{jaszkiewicz2004comparative}.  It is based on the idea of an algorithm
proposed by Ishibuchi and Murata \cite{ishibuchi1998multi}. Both methods choose weight vectors used in the scalarizing functions at random, but Jaszkiewicz's MOGL uses an aggressive tournament selection (instead of roulette wheel selection, used by Ishibuchi and Murata) to select very good solutions for recombination. The selection is aggressive in the sense that a relatively large number of solutions takes part in the tournament, and only the first and second best solutions (according to the current scalarizing function) are selected as parents. 

To be precise, tournament selection was proposed in the updated version of MOGLS \cite{jaszkiewicz2004comparative}, while the original version \cite{jaszkiewicz2002genetic} was using a so-called temporary population, selected in each iteration, to achieve similar aggressive selection. Since tournament selection is less time-consuming than the original mechanism, it is what we chose to use it in this paper.

MOEA/D was proposed by Zhang et al. \cite{zhang2007moea}. MOEA/D generates a finite set of uniformly distributed weight vectors defining a set of scalarizing functions; Zhang et al. interpret it as a decomposition of the MOOP into a number of single objective subproblems corresponding to particular weight vectors, giving rise to its name.

Please note, however, that the two methods do not simply boil down to an independent optimization of a number of scalarizing functions. In each iteration, the parents are selected from a common population, so a parent could have been obtained with the use of another scalarizing function defined by a weight vector different (but usually similar) to the one currently in use. In other words, solutions obtained during the optimization of a given scalarizing function help to optimize other, similar scalarizing functions. 

\subsection{Main Structure of MOGLS and MOEA/D}\label{II-1}

As mentioned above, the main structure of both MOGLS and MOEA/D is the same. It is described in Algorithm \ref{alg1}.

In each iteration of the initial phase, a weight vector is chosen and used as the basis for defining a scalarizing function. A new feasible solution is then generated and improved by a local search based on the current scalarizing function. Finally, the Pareto archive is updated with the new solution. 

A weight vector is also chosen in each iteration of the main phase, after which two solutions that are relatively good on the scalarizing function, defined by the chosen weight vector, are selected as parents. A new solution (offspring) is generated by a  recombination of the parents, and afterwards improved by a local search. At the end of each iteration, the Pareto archive is updated with the new offspring.    

\subsection{Selection of Weights}\label{II-2}
One aforementioned point of difference between MOGLS and MOEA/D is the way in which they choose their weight vector in each iteration. MOGLS draws a weight vector at random in each iteration with the algorithm proposed in \cite{jaszkiewicz2002genetic}, whereas in MOEA/D a finite set of uniformly distributed weight vectors is generated at the beginning of the algorithm. Then, in each iteration, MOEA/D chooses the next weight vector from this set. 

\subsection{Selection of Parents}\label{II-3}
Another different element in MOGLS and MOEA/D is the
parents selection. Please note that by selection of parents we
mean the whole process influencing the final choice of parents.
This process includes the choice of the set (population) from
which the parents are selected, the mechanism for updating
this population and the mechanism for final selection of
parents from this population.

In MOGLS, two parents are chosen by a tournament selection from the whole Pareto archive. In more detail, in each iteration a sample $T$ is drawn at random from the Pareto archive. Then, two solutions (parents) which are the best on the current scalarizing function are selected from $T$. The size of this tournament is determined in a way which assures that the two selected solutions have a specified expected rank \cite{jaszkiewicz2004comparative}, by which we mean the position of the solution in an order induced by the current scalarizing function $s$ in the whole Pareto archive, where the best solution for $s$ having a rank of 1. Assume that $T$ solutions are selected for the tournament. As shown in \cite{jaszkiewicz2004comparative}, the expected rank $Er$ of the best solution among the sample of $T$ randomly selected solutions is approximated well by:
\[ Er \approx \frac{3|\widehat{\mathcal{X}}_E|}{2T} \]
So, the larger the size of the tournament sample compared to the size of the Pareto archive, the better the solutions selected through the tournament. 
Note that in this paper we introduce a slight modification of MOGLS, namely we select parents from the Pareto archive, while the original versions of MOGLS \cite{jaszkiewicz2002genetic}\cite{jaszkiewicz2004comparative} used an additional population. This modification in fact simplifies the algorithm, and in preliminary experiments we observed that this simplified version performs at least as well as the original. For some problems, however, the Pareto archive may contain too few solutions. In such cases, it may be beneficial to keep an additional population (e.g., composed of solutions removed from the Pareto archive in recent iterations) and to select parents from both this population and the Pareto archive. No such additional population was found to be necessary for the problems and instances used in this paper. 

In MOEA/D, a single solution is kept for each of the uniformly distributed weight vectors. Furthermore, a neighborhood relation among uniformly distributed weight vectors (and thus corresponding sub-problems) is defined based on their Euclidean distance in the weights space. More precisely, a neighborhood of a given weight vector is composed of a number of its closest weight vectors. In the original version of MOEA/D, the two parents are selected from a subset of solutions corresponding to the neighbor weight vectors. In this paper, we use a newer version of MOEA/D which is inspired by the idea of algorithm was proposed in \cite{Zhang2009} in which  parents are selected from either the set of  solutions corresponding to all sub-problems or the subset of solutions corresponding to the neighbor sub-problems based on a probability. Furthermore, the new version of MOEA/D updates a specific number of solutions in each iteration while in the original version all solutions of the neighborhood sub-problems are updated.

Zhang et al. \cite{zhang2007moea} argue that this way selecting parents is faster than the mechanism used in MOGLS. Though this is indeed true, in MOGLS and MOEA/D the vast majority of CPU time is spent in local search, while the time needed to select parents is practically negligible.

Please note that the expected rank in MOGLS plays a role similar to the size of the neighborhood in MOEA/D. The lower the expected rank, and the lower the size of the neighborhood, the better the solutions for the current scalarizing function selected for recombination are on average.
\section{Uniform MOGLS}\label{III}
As stated above, our goal is to experimentally assess which of the two elements differentiating MOGLS and MOEA/D has greater influence on performance and which versions of these elements allow for better results. However, If
we observe some differences in the performance of the two
methods we would not know, however, which of the two
different elements is the main source of these differences. Therefore, we propose an intermediate method, called Uniform MOGLS (UMOGLS), which is different in just one element from both MOGLS and MOEA/D. UMOGLS selects weight vectors from a set of uniformly distributed weight vectors similarly to MOEA/D, but it chooses the solutions for recombination in the same way as MOGLS.

\begin{algorithm}[!h]
\caption{\texttt{Main Structure}}\label{algoU2PPLS}
\begin{algorithmic}\label{alg1}
\STATE Parameter $\downarrow$: Number of Initial Solutions $K$
\STATE Parameter $\downarrow$: Stopping criterion
\STATE Parameter $\uparrow$: Pareto archive $\widehat{\mathcal{A}}$ 
\vspace*{1\baselineskip}
\STATE -$\,$-$|$ \textbf{Initial Phase }
\STATE $\widehat{\mathcal{A}}$:= $\emptyset$
\FOR{($i=0;i<K;i++$)}
\STATE \texttt{GetWeight}(${\Lambda} \uparrow$)
\STATE {Construct a new feasible solution $X$}
\STATE
\texttt {LocalSearch($s(z,{\Lambda})\downarrow, X\updownarrow)$}
\STATE \texttt{Update($X\downarrow,\widehat{\mathcal{A}}\updownarrow$)}
\ENDFOR
\vspace*{1\baselineskip}
\STATE -$\,$-$|$ \textbf{Main
Phase }
\REPEAT
\STATE \texttt{GetWeight}(${\Lambda} \uparrow$)
\STATE \texttt{GetParents}($s(z,{\Lambda})\downarrow, X_1\uparrow, X_2\uparrow$)
\STATE \texttt{Recombine}(${X_1}\downarrow,{X_2} \downarrow,{X_3}\uparrow$ )
\STATE
\texttt {LocalSearch($s(z,{\Lambda})\downarrow, X_3\updownarrow)$}
\STATE \texttt{Update($X_3\downarrow,\widehat{\mathcal{A}}\updownarrow$)}

\UNTIL{the stopping criterion is met}

\end{algorithmic}
\end{algorithm}
\begin{algorithm}[!h]
\caption{\texttt{GetWeight(MOGLS)}}\label{algoU2PPLS}
\begin{algorithmic}\label{alg2}
\STATE Parameter $\uparrow$: Weight vector  ${\Lambda}$
\vspace*{1\baselineskip}
\STATE{Draw at random a normalized weight vector  ${\Lambda}$}

\end{algorithmic}
\end{algorithm}

\begin{algorithm}[!h]
\caption{\texttt{GetWeight(MOEA/D, UMOGLS)}}\label{algoU2PPLS}
\begin{algorithmic}\label{alg3}
\STATE Parameter $\uparrow$: Weight vector  ${\Lambda}$
\vspace*{1\baselineskip}
\STATE
\IF{{(it is the first time)}}
\STATE
  Generate $N$ uniformly\_distributed weight vectors: $\Psi_s=\{\Lambda^1,...,\Lambda^N\}$
\ENDIF
\STATE{Take the next weight vector from $\Psi_s$. If there is no next weight vector, use the first one}

\end{algorithmic}
\end{algorithm}

\begin{algorithm}[!h]
\caption{\texttt{GetParents (MOGLS, UMOGLS)}}\label{algoU2PPLS}
\begin{algorithmic}\label{alg4}
\STATE Parameter $\downarrow$: Scalarizing function $s_1(z,\Lambda)$  
\STATE Parameter $\downarrow$: Pareto archive $\widehat{\mathcal{A}}$  
\STATE Parameter $\downarrow$: Size of tournament $T$
\STATE Parameter $\uparrow$: Parents  $X_1$,$X_2$
\vspace*{1\baselineskip}

\STATE{Select at random $T$ solutions from  $\widehat{\mathcal{A}}$ } for tournament
\STATE{Choose the first $X_1$ and second $X_2$ best solutions in the tournament as parents }

\end{algorithmic}
\end{algorithm}

\begin{algorithm}[!h]
\caption{\texttt{GetParents (MOEA/D)}}\label{algoU2PPLS}
\begin{algorithmic}\label{alg5}
\STATE Parameter $\downarrow$: Number of neighborhood weight vectors $N$ 
\STATE Parameter $\downarrow$: Uniformly\_distributed weight vectors $\Psi_s$
\STATE Parameter $\uparrow$: Parents  $X_1$,$X_2$
\vspace*{1\baselineskip}
\IF{{(it is the first time)}}
\STATE
for each weight vector $i$ in $\Psi_s$, let $B(i)$ = \{$N$ closest weight vectors to $i$\}
\ENDIF
\STATE{Randomly select two indexes $k$ , $l$ from $B(i)$ and use related solutions as parents}

\end{algorithmic}
\end{algorithm}
\newpage

 \section{Computational experiment} \label{IV}
In order to experimentally assess the influence of different elements in MOGLS and MOEA/D, we compare the algorithms in instances of three different multiobjective combinatorial problems, i.e., multiobjective symmetric traveling salesperson problem (TSP), traveling salesperson problem with profits, and multiobjective set covering problem. To avoid the influence of implementation details, all methods were implemented in Java, sharing as much of the code as possible.

\subsection{Multiobjective symmetric TSP} \label{IV-1}
Given $N$ cities (nodes) and the traveling costs (distances) $c^k_{i,j}$ (with $ i\neq j$) between each pair of distinct cities, the multiobjective traveling salesperson problem consists of finding a circular path visiting each city exactly once. In other words, the goal is to find a permutation $p$ of the cities that minimizes the following objectives ($j=1,...,J$):

\[``{\textrm{minimize}} \textrm{''} z_j(p) = \sum_{i=1}^{N-1} c^j_{p(i),p(i+1)} + c^j_{p(N),p(1)} \]

In this paper, we use the symmetric version of the multiobjective traveling salesperson problem (MSTSP), where: 
$ c^j_{i,l}= c^j_{l,i} \mbox{ for } 1\leq i,l \leq N$.
\subsection{Multiobjective TSP with Profit} \label{IV-2}
An extension of TSP is TSP with profit (TSPWP) \cite{feillet2005traveling}. It is formulated as follows: given the set of $N$ cities and profits associated with each city, the goal is to find a sub-tour of the cities in order to minimize the tour length and maximize the collected profit.

TSPWP is multiobjective in nature \cite{jozefowiez2008multi}. However, it is usually thought of as a single objective problem and solved by an aggregation of the two objectives \cite{feillet2005traveling}. TSPWP is a problem with heterogeneous objectives, i.e, the objectives are defined by functions of different mathematical form. 

\subsection{Multiobjective set covering problem}
The MOSCP is the problem of covering the rows of an $L$-row, $I$-column, zero-one matrix in which elements are denoted by $a_{li}$, $l=1,...,L,$ and $i=1,...,I,$ by a subset of the columns minimizing $J$ cost-type objectives \cite{jaszkiewicz2003multiple}. Defining $x_i = 1$ if column $i$ (with cost $c_i^j > 0, j =1,...,J$) is selected in the solution and $x_i=0$; otherwise, the MOSCP is 
\[\textrm{minimize}\;\;\;\; \{ z_1 = \sum_{i=1}^{I} c_i^1x_i,...,z_J = \sum_{i=1}^{J} c_i^I x_i \}\]
\[\textrm{s.t.} \;\;\;\;\;\; \sum_{i=1}^{I} a_{li}x_i \geq {1}, \;\;\;\; l=1,...,L \]
\[\;\;\;\;\;\;\; x_i \in \{0,1\},\;\;\;\; j =1,...,I.\]
\subsection{Quality indicators} \label{IV-3}
In this paper we use the following quality measures:
\begin{itemize}
\item The $R$ measure \cite{hansen1998evaluating} \cite{jaszkiewicz2002genetic}: evaluates a Pareto archive  $\widehat{\mathcal{A}}$   by the average value of the weighted Chebycheff scalarizing functions over a set of normalized weight vectors. It is calculated as follows:

\[ R(\widehat{\mathcal{A}}) = \frac{\sum_{\substack{\Lambda \in \Psi_s}}^{}\displaystyle \min_{z \in A}{s_\infty (z,z^*,\Lambda)}}{|\Psi_s|} \] 

Where $\Psi_s$ is the set of uniformly distributed weight vectors generated with the procedure described in \cite{jaszkiewicz2002genetic}. 

\item Hypervolume $(HV)$ \cite{zitzler1999evolutionary}: indicates the area in the objective space that is dominated by at least one solution of the nondominated set. $HV$ of a given Pareto archive $\widehat{\mathcal{A}}$ is the Lebesgue measure of the set $\bigcup\limits_{z \in \widehat{\mathcal{A}}} H(z, r_*)$, where $r_* \in \mathbb{R}^J$ is a reference point dominated by each point in the archive and $H(z, r_*)$ is a hypercuboid defined by points $z$ and $r_*$. 

\end{itemize}
\subsection{Adaption of the methods to MSTSP} We use the 2-opt local search with two-edge exchange move, first proposed by Croes \cite{croes1958method} and possible to apply to TSP and many related problems. It consists of testing all pairs of nonadjacent edges in the tour in order to find the best pair of edges $\langle a, b \rangle$ and $\langle c,d\rangle$, such that replacing them with edges $\langle a,c \rangle $ and $\langle b,d \rangle$ results in a shorter tour.

Since local search is the most time consuming part of each
method, we use a speed-up technique, namely candidate lists,
in 2-opt local search in the main phase of each method. This technique is able to reduce the running time significantly with only a very small degradation of the quality of the retrieved solutions \cite{lust2010speed}\cite{steiglitz1968some}. There are several ways of making the candidate lists. In this paper, we use the population of initial solutions improved by the local search to make a candidate list for each node. Specifically, the candidate list of a node $a$ contains all nodes connected to $a$ in at least one of the initial solutions. Then we just consider the pairs of edges $\langle a, b \rangle$ and $\langle c,d\rangle$ such that $c$ is in the candidate list of $a$ or $d$ is in the candidate list of $b$.

For the recombination of solutions, we use the distance preserving crossover (DPX) operator \cite{freisleben1996new}. DPX generates an offspring
by putting the edges which are common in both parents to it.
The offspring is then completed by randomly selected edges
which are not present in any of the parents. As a result, the
generated offspring has the same distance (measured by the
number of different edges) to both of its parents..

In preliminary experiments with MSTSP, we observed that the best results are obtained with linear scalaring functions. We, therefore, used this type of function for this problem, similarly to \cite{jaszkiewicz2002genetic} and \cite{zhang2007moea}. 

\subsection{Adaption of the methods to TSPWP}

In TSPWP, we use a local search which performs moves of four different types. In each iteration of the local search, all moves of every type are tested, and the best move is performed.
\begin{itemize}
\item Edge exchange: This move works exactly like the two edge exchange used in MSTSP. It can change the length objective but cannot change profit, so other types of
moves are necessary.
\item Node insertion: in this move, a node which is not present in the current tour is added to the best position in the tour.
\item Node deletion: in this move, a node is deleted from the tour. 
\item Node exchange: in this move, a node present in the tour is exchanged for another node that is not present in the tour. 
\end{itemize}
\begin{algorithm}[!h]
\caption{\texttt{DPX operator for TSPWP}}\label{ExtendedDPX}
\begin{algorithmic}\label{alg6}
\STATE Parameter $\downarrow$: The parents $X_1$,$X_2$
\STATE Parameter $\uparrow$: The offspring  $X_3$

\vspace*{1\baselineskip}
\STATE {$comSet$ := set of common edges and nodes in $X_1$ and $X_2$}
\STATE {$remSet$ := set of remaining nodes not present in $comSet$}
\STATE {$fragmentSet$ := $comSet$}
\STATE {$expectedNumberOfNodes$ := average number of nodes in $X_1$ and $X_2$}
\STATE $addProbability := \frac{expectedNumberOfNodes-|comSet|}{|remSet|}$
\FOR {\textbf{each} node $a \in remSet$}
\STATE {add $a$ to $fragmentSet$ with probability $addProbability$}
\ENDFOR
\STATE Combine the edges and nodes from $fragmentSet$ randomly creating a circular path
\end{algorithmic}
\end{algorithm}
For the recombination operator, we use an extended version of the DPX operator, in which we collect both common nodes and common edges between two parents. We then randomly add a few remaining nodes to obtain the expected number of nodes, equal to the average number of nodes in the parents. The fragments (edges and nodes) are then combined randomly, creating a circular path. The extended version of the DPX operator is explained in algorithm \ref{ExtendedDPX}.

In preliminary experiments with TSPWP we observed that the best results are achieved through mixed scalaring functions with weights $0,999$ for the Chebycheff scalarizing function, and $0,001$ for the linear scalarizing function. Thus, this type of function was employed for this problem. 

Since the two objectives may have very different ranges, we normalize their values using certain approximate ranges of the objectives in the Pareto front. In more detail, the approximate ranges were retrieved, at the beginning of each method, by running local search with two scalarizing functions with weight vectors (0.999, 0.001) and (0.001, 0.999). 

\subsection{Adaption of the methods to MOSCP}
In MOSCP, the local search is performed based on a neighborhood operator, which is guided by a scalarizing function and defined as follows \cite{jaszkiewicz2004comparative}:
first, a randomly selected column is removed from the solution. This leads to an unfeasible solution. The solution is then repaired in a greedy manner by inserting columns with the lowest ratio of:\\
\[\frac{\textnormal{scalarizing value decline caused by insertion of the column}}{\textnormal{the number of uncovered rows covered by the column}}\]\\
The column removed in the first step is not considered by the greedy procedure, therefore, the neighborhood operator always produces a new solution. The whole neighborhood of the current solution is tested and the best local move is performed.
The recombination operator is also based on the distance-preserving crossover idea. An offspring is generated as follows: first, all columns common to both  parents are inserted into the offspring. Then, the columns which appear in only one of the parents are inserted into the offspring with $50\%$ probability. Since this procedure cannot guarantee covering all rows, lastly, all uncovered rows are covered with columns selected randomly.

\subsection{Experiment design}
We present the average values of the quality indicators over 10 executions for each method and each instance. We compare four methods: Multiobjective Multiple Start Local Search (MOMSLS), MOGLS, UMOGLS, and MOEA/D for MSTSP, TSPWP, and MOSCP. To avoid influence of implementation 
differences, all methods were implemented in Java from scratch sharing as much of the code as possible.
 
MOMSLS is a simple method employing multiple runs of a local search. Each run  starts with a random initial solution and uses a scalarizing function with a random weight vector. In other words, MOMSLS is similar to the initial phases of MOGLS and MOEA/D, and is therefore a natural reference to MOGLS and MOEA/D. The use of recombination in these methods should ensure a better performance than MOMSLS.

Two different types of instances of MTSP have been used:
\begin{itemize}
\item Euclidean instances: in this group, the distances between the edges correspond to the Euclidean distances between points randomly located in a plane with uniform distribution. Euclidean and Kro instances are included into this group.

\item Cluster instances: the points are randomly clustered in a plane and the distances between points correspond to their Euclidean distance.
\end{itemize}
For two- and  three-objective instances of MSTSP, we use the instances which were proposed in Lust's library\footnote{\url{https://www-desir.lip6.fr/~lustt/Research.html##MOTSP} \label{fotnote}}\cite{Lust2010}. 
As mentioned in section \ref{IV-2}, in TSPWP the first objective is the length of the tour, while the second objective is the collected profit. In our experiment, the first objective comes from either Euclidean or Cluster instances, and the profits are generated randomly from a uniform distribution in a given range. 

We use bi-objective instances of MOSCP from Lust's library\footnote{\url{https://www-desir.lip6.fr/~lustt/Research.html##MOSCP} \label{fotnote}}\cite{Lust2014}. We generated three-objective instances of MOSCP by combining two bi-objective instances. Two objectives came from the first instance, and the third objective was the first from the second instance. The instances are available from the authors upon request.

For a fair comparison, the number of weights in MOEA/D and UMOGLS, and the number of initial solutions in MOGLS, were set the same way in all methods. The number of iterations was also the same in all methods. As a consequence, the same total numbers of local search runs and recombinations were performed in all methods. The number of iterations in MOMSLS was also the same as the number of iterations in MOGLS, UMOGLS, and MOEA/D. By one iteration we mean one run of local search (MOMSLS and initial phases of other methods), or one recombination and one run of local search (main phases of MOGLS, UMOGLS, and MOEA/D). 

The parameters of each method were set experimentally, based on the best choice principle. The parameter setting for particular instances is listed in Table \ref{TP}. The size of the neighborhood in MOEA/D is set to 20, the probability for choosing parents from a subset of solutions corresponding to the neighbor weight vectors is set to 0.9 and the number of solution which will be updated in each iteration is set to 2. Also the expected rank value in MOGLS and UMOGLS is set to 10, 5, 4 for instances KroAB100, ClusterAB300, and EuclideanAB500 respectively. And it is set to 10 and 8 for instances KroABC100 and ClusterABC300 respectively. For all other instances, the expected rank is set to 10.

The number of weight vectors used in the quality measure \(R\) was set to 1000 for all bi-objective instances, and to 7562 for all three-objective instances. The reference points are defined by the minimum values of each objective in the reference sets. 
\begin{table}[ht]
\caption{Parameter setting} \label{TP}
 \centering
  \scalebox{0.75}{
 \begin{tabular}{ccccc}
 \hline \scriptsize Problem & \scriptsize Number of Generations &  \scriptsize  Number of Weight Vectors  \\ \hline
        {\scriptsize MSTSP-2obj}
   & \texttt{\scriptsize $50$}  & \texttt{\scriptsize $101$} \\ [0.05ex]    \hline
      {\scriptsize MSTSP-3obj}
   & \texttt{\scriptsize $5$}  & \texttt{\scriptsize $3403$} \\ [0.05ex]    \hline
      {\scriptsize TSPWP}
   & \texttt{\scriptsize $17$}  & \texttt{\scriptsize $301$} \\ [0.05ex]    \hline
       {\scriptsize MOSCP-2obj}
   & \texttt{\scriptsize $17$}  & \texttt{\scriptsize $301$} \\ [0.05ex]    \hline
      {\scriptsize MOSCP-3obj}
   & \texttt{\scriptsize $5$}  & \texttt{\scriptsize $3403$} \\ [0.05ex]    \hline
    
 \end{tabular}}
 \end{table}
\begin{table}[h]
\caption{Results for 2-obj instances of MSTSP} \label{T1}
\centering
\scalebox{0.76}{
\begin{tabular}{llllll} 
\hline \scriptsize Instance & \scriptsize Quality & \scriptsize MOMSLS &        \scriptsize  MOGLS & \scriptsize UMOGLS & \scriptsize MOEA/D \\ \hline
 \multirow{1}{20mm}{\scriptsize KroAB100}
    & \texttt{\scriptsize $R$}  & \scriptsize 10765.39(7.92) & \scriptsize 10408.17(11.27) & \scriptsize \textbf{10405.71}(8.81) & \scriptsize 10508.75(26.52) \\ [0.01ex]
    
       & \texttt{\scriptsize $HV$}  & \scriptsize 21.71E+09

(5.72E+06)

) & \scriptsize 21915.43E+06
(3.71E+06) & \scriptsize \textbf{21915.56E+06}
(4.32E+06) & \scriptsize 21.85E+09

(7.94E+06) \\ [-0.2ex] \hline

\multirow{1}{20mm}{\scriptsize ClusterAB300}
    & \texttt{\scriptsize $R$}  & \scriptsize 27187.44(17.24) & \scriptsize 26221.41(31.70) & \scriptsize \textbf{26212.33}(47.61) & \scriptsize 26612.22(26.04) \\ [0.01ex]
    
       & \texttt{\scriptsize $HV$}  & \scriptsize 211.35E+09

(2.20E+07

) & \scriptsize \textbf{2125.61E+08}
(5.63E+07

) & \scriptsize 2125.42E+08
(6.07E+07

) & \scriptsize 211.88E+09
(2.77E+07) \\ [0.01ex] \hline

\multirow{1}{20mm}{\scriptsize EuclideanAB500}
    & \texttt{\scriptsize $R$}  & \scriptsize 51015.59(18.71) & \scriptsize \textbf{49117.52}(55.62) & \scriptsize 49119.42(45.35)& \scriptsize 49921 .83(49.81) \\ [0.01ex]
    
       & \texttt{\scriptsize $HV$}  & \scriptsize 5.79E+11
(3.51E+07
) & \scriptsize 583.55E+09

(1.39E+08
) & \scriptsize \textbf{583.61E+09
}
(1.13E+08
) & \scriptsize 5.81E+11

(9.15E+07
) \\ [0.01ex] \hline   
\
\end{tabular}
}
\end{table}

\begin{table}[h]
\caption{Results for 3-obj instances of MSTSP}\label{T2}
\scalebox{0.76}{
\begin{tabular}{llllll}
\hline \scriptsize Instance & \scriptsize Quality & \scriptsize MOMSLS &        \scriptsize  MOGLS & \scriptsize UMOGLS & \scriptsize MOEA/D \\ \hline
  \multirow{1}{20mm}{\scriptsize KroABC100}
    & \texttt{\scriptsize $R$}  & \scriptsize 12708.28
(4.84
) & \scriptsize 12358.69
(5.29
) & \scriptsize \textbf{12353.63}
(5.34
) & \scriptsize 12454.55
(4.2
) \\ [0.01ex]
    
       & \texttt{\scriptsize $HV$}  & \scriptsize 3.57E+15
(5.97E+11

) & \scriptsize \textbf{3633.83E+12}

(7.87E+11

) & \scriptsize 3633.71E+12

(7.21E+11

) & \scriptsize 3.61E+15

(4.55E+11

) \\ [-0.2ex] \hline

\multirow{1}{20mm}{\scriptsize  ClusterABC300}
    & \texttt{\scriptsize $R$}  & \scriptsize 17026.97 

(3,72) & \scriptsize 16723.42

(3,05
) & \scriptsize \textbf{16701.25 
}

(5.3

) & \scriptsize 16837.89

(4,23
) \\ [0.01ex]
    
       & \texttt{\scriptsize $HV$}  & \scriptsize 11.27E+16

(1.28E+13

) & \scriptsize 1130.35E+14

(1.86E+13

) & \scriptsize \textbf{1130.59E+14 }

(6.75E+12

) & \scriptsize 11.29E+16

(8.96E+12

) \\ [0.01ex] \hline  
\end{tabular}
}
\end{table}
\begin{table}[h]
\caption{Results for instances of TSPWP}\label{T3}
\centering
\scalebox{0.76}{
\begin{tabular}{llllllll}
\hline \scriptsize Instance & \scriptsize Quality & \scriptsize MOMSLS &  \scriptsize  MOGLS & \scriptsize UMOGLS  & \scriptsize MOEA/D \\ \hline
  \multirow{1}{20mm}{\scriptsize KroAProfit100}
    & \texttt{\scriptsize $R$}  & \scriptsize 0.16
(3.59E-04

) & \scriptsize \textbf{0.1587}
(1.20E-04

) & \scriptsize 0.1589

(1.81E-04

) & \scriptsize 0.159

(1.56E-04

)  \\ [0.01ex]
    
      & \texttt{\scriptsize $HV$}  & \scriptsize 4.42E+08

(1.31E+06) & \scriptsize \textbf{467.97E+06}

(4.47E+05) & \scriptsize 467.25E+06

(5.86E+05) & \scriptsize 46.44E+07

(7.88E+05)  \\ [0.01ex] \hline
        \multirow{1}{20mm}{\scriptsize ClusterAProfit300}
   & \texttt{\scriptsize $R$}  & \scriptsize 0.156
(3.41E-04

) & \scriptsize \textbf{0.1445}
(2.47E-04

) & \scriptsize 0.1446
(3.11E-04

) & \scriptsize 0.151
(3.34E-04

)   \\ [0.01ex]
    
     & \texttt{\scriptsize $HV$}  & \scriptsize 3.01E+09

(6.47E+06

) & \scriptsize 33.25E+08

(6.28E+06) & \scriptsize  \textbf{33.37E+08}

(1.14E+07

) & \scriptsize 3.22E+09

(8.39E+06

)    \\ [0.01ex] \hline
      
\end{tabular}
}
\end{table}

\begin{table}[h]
\caption{Comparison between methods (2-obj instances of MOSCP).}\label{T3}
\centering
\scalebox{0.76}{
\begin{tabular}{llllllll}
\hline \scriptsize Instance & \scriptsize Quality & \scriptsize MOMSLS &  \scriptsize  MOGLS & \scriptsize UMOGLS  & \scriptsize MOEA/D \\ \hline
  \multirow{1}{20mm}{\scriptsize 2scp41A}
    & \texttt{\scriptsize $R$}  & \scriptsize 180.77
(0.28
) & \scriptsize 179.19
(0.09
) & \scriptsize \textbf{179.16}
(0.03
) & \scriptsize 179.35
(0.18
)  \\ [0.01ex]
    
      & \texttt{\scriptsize $HV$}  & \scriptsize 38.18E+05

(5.12E+03

) & \scriptsize \textbf{3840.54E+03}

(8.39E+02

) & \scriptsize 3840.15E+03

(2.31E+02

) & \scriptsize 38.37E+05

(3.23E+03

)  \\ [0.01ex] \hline
      
        \multirow{1}{20mm}{\scriptsize 2scp61A}
   & \texttt{\scriptsize $R$}  & \scriptsize 549.09
(0.78
) & \scriptsize \textbf{537.80}
(0.22
) & \scriptsize 537.88
(0.35
) & \scriptsize 538.68
(0.34
)   \\ [0.01ex]
    
     & \texttt{\scriptsize $HV$}  & \scriptsize 66.61E+06

(2.49E+04

) & \scriptsize 67034.46E+03

(3.08E+04

) & \scriptsize \textbf{67034.62E+03}

(3.05E+04
) & \scriptsize 66.96E+06
(3.36E+04)  \\ [0.01ex] \hline
       \multirow{1}{20mm}{\scriptsize 2scp81A}
   & \texttt{\scriptsize $R$}  & \scriptsize 1077.91

(1.35

) & \scriptsize 1050.83

(0.66
) & \scriptsize \textbf{1050.70}
(0.36
) & \scriptsize 1052.75
(0.69
)   \\ [0.01ex]
    
     & \texttt{\scriptsize $HV$}  & \scriptsize  16.71E+07

(4.79E+04

) & \scriptsize \textbf{168.21E+06}

(5.13E+04

) & \scriptsize 168.19E+06

(3.46E+04

) & \scriptsize 168.09E+06

(4.19E+04

)   \\ [0.01ex] \hline

\end{tabular}}
\end{table}
\begin{table}[ht]
\caption{Comparison between methods (3-obj instances of MOSCP).}\label{T3}
\centering
 \scalebox{0.76}{
\begin{tabular}{llllllll}
\hline \scriptsize Instance & \scriptsize Quality & \scriptsize MOMSLS &  \scriptsize  MOGLS & \scriptsize UMOGLS  & \scriptsize MOEA/D \\ \hline
  \multirow{1}{20mm}{\scriptsize 3scp41A}
    & \texttt{\scriptsize $R$}  & \scriptsize 184.14
(0.27
) & \scriptsize \textbf{180.24}
(0.09
) & \scriptsize 180.24
(0.14
) & \scriptsize 180.86
(0.12
)  \\ [0.01ex]
    
      & \texttt{\scriptsize $HV$}  & \scriptsize 14.61E+09

(7.94E+06) & \scriptsize \textbf{14738.61E+06
}

(2.26E+06) & \scriptsize 14738.41E+06

(3.17E+06) & \scriptsize 147.15E+08

(3.06E+06)   \\ [0.01ex] \hline
        \multirow{1}{20mm}{\scriptsize 3scp61A}
   & \texttt{\scriptsize $R$}  & \scriptsize 707.72
(1.38
) & \scriptsize \textbf{665.06}
(0.63
) & \scriptsize 665.88
(0.77
) & \scriptsize 673.27
(1.23
)   \\ [0.01ex]
    
     & \texttt{\scriptsize $HV$}  & \scriptsize 8.31E+11

(7.85E+08

) & \scriptsize \textbf{855.48E+09
}

(2.97E+08

) & \scriptsize 855.28E+09

(3.96E+08

) & \scriptsize 85.04E+10

(5.06E+08

)   \\ [0.01ex] \hline
       \multirow{1}{20mm}{\scriptsize 3scp81A}
   & \texttt{\scriptsize $R$}  & \scriptsize 1362.97
(1.58
) & \scriptsize \textbf{1271.18}
(2.49
) & \scriptsize 1272.83
(1.99
) & \scriptsize 1298.7
(1.59
)   \\ [0.01ex]
    
     & \texttt{\scriptsize $HV$}  & \scriptsize 3.67E+12
(3.69E+09

) & \scriptsize \textbf{38.30E+11}

(3.40E+09

) & \scriptsize 38.29E+11

(2.34E+09

) & \scriptsize 3.77E+12

(2.38E+09

)   \\ [0.01ex] \hline
\end{tabular}}
\end{table}

 \begin{table}[h]
 \caption{Influence of number of Weight vector on instance of 2scp81A} \label{TW}
 \centering
  \scalebox{0.76}{
 \begin{tabular}{lllll}
 \hline \scriptsize Number of Weight vector & \scriptsize Quality &  \scriptsize  MOGLS  &  \scriptsize UMOGLS  & \scriptsize MOEA/D \\ \hline
        \multirow{1}{30mm}{\scriptsize 101 Weight vectors}
   & \texttt{\scriptsize $R$}  & \scriptsize 1050.69
(0.37
) & \scriptsize 1051.77
(0.53
) & \scriptsize 1054.60
(1.58
)  \\ [0.05ex]
    
     & \texttt{\scriptsize $HV$}  & \scriptsize 186.79E+06
(3.95E+04
) & \scriptsize  186.58E+06
(4.19E+04
) & \scriptsize 186.39E+0.6
(7.15E+04
)   \\ [0.05ex] \hline
     \multirow{1}{30mm}{\scriptsize  201 Weight vectors}
    & \texttt{\scriptsize $R$}  & \scriptsize\ 1050.92
 
(0.38) & \scriptsize 1051.19
 
(0.48

) & \scriptsize 1052.97
(0.32)   \\ [0.05ex]
    
   & \texttt{\scriptsize $HV$}  & \scriptsize 1867.82E+05
(4.03E+04

) & \scriptsize 1867.41E+05
(4.08E+04

) & \scriptsize 186.63E+06
(2.38E+04
)   \\ [0.05ex] \hline
       \multirow{1}{30mm}{\scriptsize 301 Weight vectors}
    & \texttt{\scriptsize $R$}  & \scriptsize  1050.83(0.66) & \scriptsize 1050.70(0.36) & \scriptsize  1052.75((0.69)  \\ [0.05ex]
    
   & \texttt{\scriptsize $HV$}  & \scriptsize 168.21E+06 
 ((5.13E+04 
) & \scriptsize  168.19E+06
 (3.46E+04
) & \scriptsize 168.09E+06
(4.19E+04 
)   \\ [0.05ex] \hline
 \end{tabular}}
 \end{table}

\section{Results and discussion} \label{V}
Tables \ref{T1}-\ref{T3} present the mean and the standard deviation of quality indicators values obtained by MOMSLS, MOGLS, UMOGLS, and MOEA/D on MSTSP, TSPWP, and MOSCP instances. We do not report running times as they were very similar for MOGLS, UMOGLS, and MOEA/D due to the same number of iterations. Running times for MOMSLS were always higher than for other methods, as local search starting from random solutions takes more time than that starting from an offspring created by recombination.

The main observations are:
\begin{itemize}
\item In all cases, MOMSLS performs worst compared to all other methods. This proves that the use of recombination to generate a starting solution for local search heavily influences the performance of MOEAs.

\item In all test instances, the best results were attained by either MOGLS or UMOGLS. It is also apparent that MOGLS and UMOGLS work very similarly.

\item MOEA/D never obtains the best values among other tested methods, and its results are substantially worse compared to MOGLS and UMOGLS. 

\item In order to test the statistical significance of the differences, we performed the non-parametric Wilcoxon signed-rank test, with a level of risk equaling 5. We found, in all cases, that MOEA/D was significantly worse than MOGLS and UMOGLS. The two latter methods did not differ in any cases except the ClusterABC300 instance of MSTSP, for which UMOGLS was slightly better. As can be seen in Table \ref{T2}, the difference between UMOGLS and MOGLS in this instance was, however, much smaller than the difference between MOEA/D and any other method.
\end{itemize}
The main observation from the whole experiment is that the main design element which influences the performance of scalarizing functions-based multiobjective evolutionary algorithms is the choice of the mechanism for parent selection. MOEA/D and UMOGLS differ only in this mechanism, and their performance varies significantly. 

Furthermore, since UMOGLS performs better than MOEA/D, and the two methods differ only in parent selection, we may conclude that the selection mechanism used in UMOGLS (and MOGLS) is a better choice for the problems considered in this paper. In our opinion this is mainly due to the fact that, in MOGLS and UMOGLS, parents are selected from a larger population of solutions, i.e., from the Pareto archive, while in MOEA/D the population is bounded by the predefined number of weight vectors. Selecting parents from the Pareto archive assures better diversity and allows avoiding a premature convergence in MOGLS and UMOGLS.

On the other hand, since UMOGLS and MOGLS perform very similarly, we may conclude that selecting vector weight either at random or from a set of uniformly distributed weight vectors does not substantially influence algorithm performance.

It is worth mentioning, though, that the number of weight
vectors should be large enough to ensure that UMOGLS works
comparably to MOGLS. To illustrate this principle, in table \ref{TW} we
present the results for instance 2scp81A with 101, 201,
and 301 weight vectors obtained with MOGLS, UMOGLS,
and MOEA/D with a constant total number of iterations. Please
note that, in the case of MOGLS, the number of weight
vectors influences only the number of initial solutions. These
results show that, with 101 and 201 weight vectors, UMOGLS
is worse than MOGLS and, at the same time,
significantly better than MOEA/D. However, for 301 weight
vectors, MOGLS and UMOGLS are alike but at the
same time significantly better than MOEA/D.
Also note that the performance of MOEA/D improves
with the growing number of weight vectors, probably due
to an increasing population size from which parents are
selected. The performance of MOEA/D still improves
more slowly than the performance of UMOGLS.\\
Observe that MOGLS and MOEA/D algorithms have additionally been experimentally compared in \cite{zhang2007moea} regarding the multiobjective knapsack problem. In that paper MOEA/D performed better than MOGLS. In our opinion, one source of the differences in results presented here could be the fact that in \cite{zhang2007moea} a considerably different approach to the adaptation of MOEAs to the specific problem was used. This adaptation did not use any local search, instead a simple greedy repair method was applied after recombination. This repair method, in general, makes smaller changes to the offspring than the local search. Furthermore, in \cite{zhang2007moea} an older version of MOGLS was used, one in which parents were selected from a temporary population \cite{jaszkiewicz2002genetic}.

\section{Conclusions} \label{vi}
We have presented an experimental comparison of MOGLS and MOEA/D algorithms on three different multiobjective combinatorial problems, i.e., the multiobjective symmetric traveling salesperson problem, traveling salesperson problem with profits, and multiobjective set covering problem. In this comparison we focused not only on the overall evaluation of the algorithms, but we also tried to identify the main design elements influencing their performance. To our knowledge, this is the first such systematic comparison of these algorithms. Such analysis provides deeper insight into the sources of variation between different methods.

Our results indicate that the main factor influencing the performance of  algorithms is the selection of parents. 
However, when the most optimal parameter settings were applied, the choice of parents with tournament selection used in MOGLS and UMOGLS performed better on all three problems used in the experiment. 

We have also proposed a slight modification of MOGLS in this paper, a simplification in fact, in which parents are selected from the Pareto archive. In other words, we do not use any additional population of solutions.

We have also proposed a new method called Uniform MOGLS, which uses predefined weight vectors akin to MOEA/D and tournament selection akin to MOGLS. This method performs very similarly to MOGLS with random weight vectors if the number of predefined weight vectors is sufficiently large. This shows that the mechanism for weight vector selection has little influence on the performance of the MOAEs if sufficient coverage of weight space is assured. 

Similar results have been obtained for three different multiobjective combinatorial problems. Without a doubt, further computational studies on other combinatorial and continuous problems, including problems with higher numbers of objectives, would be beneficial to assess whether the same pattern holds in other cases.

Recently Zhang et al. proposed \cite{li2015interrelationship} certain extended versions of MOEA/D in which the selection of solutions for the current population is performed differently, explicitly taking into account both the quality and diversity of solutions associated with particular weight vectors (sub-problems). Thus, an interesting direction for further research would be to compare MOGLS/UMOGLS to these new versions of MOEA/D.

\section*{Acknowledgment}
This research was supported by the the Polish National Science Center, grant no. UMO-2013/11/B/ST6/01075.\\
We also would like to thank Dr. Manuel López-Ibáñez (University of Manchester, UK) 
for his helpful comments on this work.

\section*{References}

\bibliography{mybibfile}

\end{document}